# An application-oriented terminology evaluation: the case of back-of-the book indexes

**Touria Aït El Mekki[1] and Adeline Nazarenko[2]**

[1] LERIA
touria at info.univ-angers.fr
[2] LIPN UMR7030
Université Paris 13 & CNRS,
99 avneue J.B. Clément 93430 Villetaneuse, FRANCE
Adeline Nazarenko at lipn.univ-paris13.fr

**Abstract**
This paper addresses the problem of computational terminology evaluation not per se but in a specific application context. This paper describes the evaluation procedure that has been used to assess the validity of our overall indexing approach and the quality of the IndDoc indexing tool. Even if user-oriented extended evaluation is irreplaceable, we argue that early evaluations are possible and they are useful for development guidance.

## 1. Introduction

Back-of-the-book indexes are precious information retrieval devices that offer an easy way to locate a given piece of information in a large document and to navigate through that document. Unfortunately, indexes are expensive to produce, because indexing remains mainly manual. Modern word processing or indexing tools provide a technical assistance but do not address the index content and information selection problem. The professional indexing tools (Sonar BookEnds, IndexingOnline, Cindex, for instance) only slightly rely on the analysis of the document content.

Arguing that computational terminology is now able to give further assistance, we have designed a new indexing method, which exploits terminological tools to facilitate the indexing task. From the analysis of the document text, our IndDoc system automatically builds an index draft that is then validated by an indexer through a dedicated interface. The resulting index is a terminological network which nodes correspond to the index entries associated with page numbers.

When developing such an innovative method, which cannot be directly compared with existing ones, one has to think of how it can be evaluated. The general approach must be validated as soon as possible, *i.e* without waiting that a user set can test an operational system. This paper addresses this preliminary evaluation problem. Even if the indexing task remains difficult to evaluate, we describe the method that we designed to nevertheless assess the quality of our approach towards indexing.

The first two sections describe what is a back-of-the-book index and present the IndDoc overall method and architecture. The section 4 explains the evaluation difficulties that one has too face when evaluating indexing tools. Our evaluation protocol is respectively presented and discussed in section 5.

## 2. Back-of-the book indexes

Traditionally, the index that is placed at the back of a book or document is an alphabetic list of descriptors associated with page numbers or page ranges. It is composed of two parts: a nomenclature and a list of references (see Figure 1).

The nomenclature is a list of descriptors, the index entries, that give access to the document content. Some index nomenclatures are structured and present explicit semantic relations between descriptors. This structure is usually mainly hierarchical. The specific descriptors are presented as sub-entries of entries that correspond to more generic descriptors (see *knowledge* and *knowledge representation* on Figure 1). Some indexes also have synonymy relations, variations (the expanded form of *AI*) or more generally association links (often *called see* or *see also*).

An index is therefore composed of a terminological network (the nomenclature made of descriptors and terminological relations) and we developed a tool to automatically produce a book index out of the docunpent terminological analysis.

| Nomenclature | References |
|---|---|
| Acquisition | 7, 21, 78-81, 250 |
| AI see Artificial Intelligence | |
| Artificial Intelligence | 43, 97, 134 |
| Knowledge | **26-32**, 76-77, 89, **211-215**, 228 |
|    Acquisition (see also Acquisition) | |
|    Representation | 25-29, 80, **132-136**, 250 |

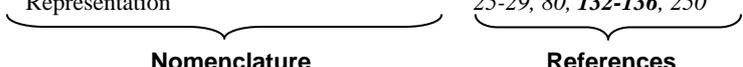

Figure 1: Index example

## 3. IndDoc indexing method and architecture

Our indexing approach is based on the last decade results in computational terminology and more generally in natural language processing. The architecture of our system is presented on Figure 2.

The terminological analysis is itself composed of two steps, the term and relation extraction respectively. For the experiment reported here, we exploited Lexter (Bourrigaut et al. 1996) and Syntex (Bourigault & Fabre, 2000) for extracting terms and we developed our own relation extraction module, which combines some contextual extraction patterns extraction (Hearst, 1992, Morin, 1999, Charniak & Berland, 1999), the syntactic analysis of the terms and the projection of a synonymy dictionary (Hamon & Nazarenko, 2001).

Once a draft of the terminological network is built, IndDoc looks for the term occurrences in order to connect the nomenclature entries with the document segments (reference calculus). The next procedure aims at ranking by relevance order both the list of terms and the set of reference segments for each index entry. This ranking procedure is important, for instance, to adjust the index length to fit the editorial constraints. It also guides the validation process. This ranking is based on the frequency of terms and their repartition over the document but also on cohesion and salience factors (typographical, lexical and textual), which establish the relative importance of index descriptors and document segments as reference candidates (Aït El Mekki & Nazarenko, 2005).

The resulting index, however, is only a draft index, since all the extracted terms and relations are not well formed or relevant for a given index. An experienced indexer must manually validate the result. We developed a dedicated interface to help this validation process.

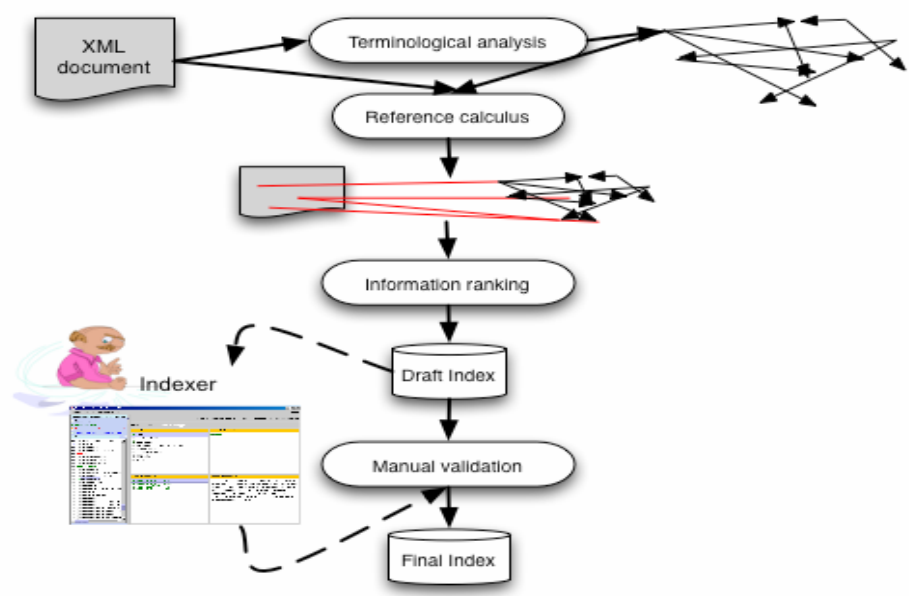

Figure 2: IndDoc architecture

## 4. Evaluation problem

When developing such a system, it is extremely important to be able to evaluate it. The goal is twofold:
- Assess the adequacy of the overall approach
- Evaluate the usefulness of the terminological tools not *per se* as it is performed in evaluation campaigns[1] and extractor comparison (Cabre *et al.*, 2001) but in the specific application context of index building.

However, evaluating IndDoc results raises two separate questions that are traditional in terminological processing. Since our indexing method is a cooperative one, it is difficult to evaluate the specific contribution of the automatic tool. It is also difficult to evaluate the quality of indexes since there is no objective reference. Two indexers do not produce the same index for a given document. The indexing guides only give general recommendations like: "Try to be as objective as possible in the choice of entries and include those entries in the index that you think a reader may want to look up. Refer only to pages where an item is discussed, not just mentioned." (Mulvany 1993). More generally it is acknowledged that indexers lack of systematic evaluation protocols (Wyman, 2005).

An additional problem comes from the fact that IndDoc is still a laboratory prototype, which cannot be easily tested by a group of users in realistic working conditions. As any system developer, we nevertheless need early evaluation elements to decide whether to pursue the development or to abandon it.

---
[1] See for instance the CESART campaign: http://www.elda.org/article137.html.

## 5. Elements of evaluation

Our indexing method should target two types of users: the indexer who builds a source index out of a draft index using a validation interface, and the reader who uses the resulting index for information localisation. However, we consider that the indexer is responsible for the adaptation of the index to the expected reader's profile. In this paper, we only evaluate the impact of the automatic indexing process on the cost and quality of the indexer's task.

The hypothesis underlying the IndDoc system development was that terminological processing would enable indexers to build richer indexes more easily than with traditional indexing tools. Really validating the above hypothesis, however, would require to have indexers testing the IndDoc system in a more systematic way and to analyze their feedback. Such a large-scale experiment cannot be set up from scratch. We need a preliminary evaluation beforehand. This is the goal of the elements of evaluation that we described here.

To get an idea of the quality of our indexing method, we compared several indexes produced for the same documents. We deliberately re-indexed documents which had been previously been published with an index. We made three types of comparisons (see the numbered bold arrows on Figure 3).

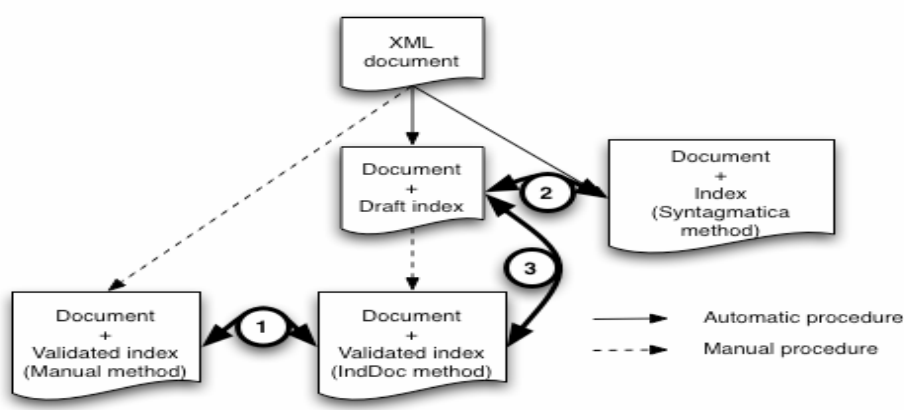

Figure 3: Schema of the evaluation protocol

In order to evaluate the added-value for an indexer to exploit an indexing tool such as IndDoc, we compare traditional indexes (traditional manual indexing) and IndDoc indexes (automatic indexing and manual validation) for the same documents[2].

We also compare the draft indexes produced by IndDoc and the equivalent draft indexes produced by existing indexing tools such as Syntactica[3], which analyses the text of the document and proposes every noun phrases as index descriptors. This aims at assessing the contribution of terminological analysis and information ranking to indexing.

We finally compare the draft indexes automatically built by IndDoc and the final indexes resulting from the indexers' validation. This comparison helps to evaluate the quality of the automatic process of IndDoc.

The experiments reported here have been performed on three different corpora, mainly focused on linguistics (LI), Artificial Intelligence (AI) and Knowledge Acquisition (KA) and the results are presented on Table 1. These figures are globally encouraging. They show that the IndDoc procedure produces much richer indexes than the traditional author's manual indexing. The size of the index considered as the number of descriptors and the proportion of relations per descriptor is significantly increased.

The second set of comparisons shows that IndDoc outperforms existing tools because it proposes some relations between descriptors and it filters out the descriptor lists (we estimate that Syntactica would produce 10 000 descriptors for KA whereas the editors who produced the final index out of IndDoc results refused to validate more than 2 000 ones).

The third type of comparison brings out contrasted results. The precision of the relation extraction is rather good (more than 65%, even though the method need to be improved) much better than the precision rates of descriptor extraction. This last result does not take the ranking into account, however. The ranked precision rates, which reflect the capacity of the system to top rank good descriptors, are much better (more than 75 %) and encouraging.

---

[2] In the reported experiments, the traditional indexes are those with which the books have been published. A different person from the original indexers, which, in this case, were the document authors, has validated the IndDoc draft indexes.

[3] The Syntactica analysis has been simulated, since Syntactica only processes English documents whereas our first IndDoc experiments were done on French documents.

|  | Monographs | | Collection |
| --- | --- | --- | --- |
|  | **LI** | **AI** | **KA** |
| Corpus size (# of words occurrences) | 42 260 | 111 371 | 122 229 |
| Existence of an original manual index | Yes | Yes | No |
| Existence of a draft index | Yes | Yes | Yes |
| Existence of an IndDoc index | Yes | Yes | Yes |
| Precision of descriptor extraction – comparison 3 | *33%* | *44%* | *71%* |
| Ranked precision of descriptor extraction – comparison 3 | *77%* | *83%* | *83%* |
| Precision of relation extraction – comparison 3 | *65 %* | *71 %* | *80%* |
| Size increase (# of descriptors) – comparison 1 | *+85%* | *+50%* | Non applicable |
| Size increase (average # of relations per descriptor) – comparison1 | *+166%* | *+300%* | Non applicable |

Table 1: Evaluation results for three different corpora. The precision figures give the proportion of relevant information in the draft index. The percentage figures show that IndDoc index is much richer than the original published indexes (the book authors acknowledged the overall quality of these large indexes).

## 6. Conclusion

Even if evaluating terminological products is known as a difficult task (man-machine cooperation, subjectivity of the quality criteria and heterogeneity of the terminological methods and goals), we showed that it is possible to evaluate the contribution of terminological tools such as term and terminological relation extraction in the context of a given application (here, back of the book indexes). This type of evaluation procedure is relatively easy to set up compared with user-based ones. It does not support a definitive assessment but it gives useful indications of the method quality prior to large experimental evaluations.